\documentclass[runningheads]{llncs}

\usepackage[mobile]{eccv}

\usepackage{eccvabbrv}

\usepackage{graphicx}
\usepackage{booktabs}

\usepackage[accsupp]{axessibility}  % Improves PDF readability for those with disabilities.

\usepackage[pagebackref,breaklinks,colorlinks,citecolor=eccvblue]{hyperref}
\usepackage{orcidlink}

\begin{document}

\title{UniDream: Unifying Diffusion Priors for Relightable Text-to-3D Generation} 

\titlerunning{UniDream: Unifying Diffusion Priors for Relightable Text-to-3D Generation}
\authorrunning{Liu et al.}

\author{
Zexiang Liu$^{1*}$, Yangguang Li$^{1*}$, Youtian Lin$^{1*}$, Xin Yu$^{2}$, Sida Peng$^{3}$,\\
Yan-Pei Cao${^1}$, Xiaojuan Qi$^{2}$, Xiaoshui Huang$^{4}$, Ding Liang${^{1\dagger}}$, Wanli Ouyang$^{4,5}$}

\institute{$^1$VAST \quad$^2$The University of Hong Kong \quad $^3$Zhejiang University \\ 
$^4$Shanghai AI Laboratory \quad $^5$The Chinese University of Hong Kong
~\\
~\\
}

\maketitle

\begin{figure*}[h]
    \centering
    \vspace{-3em}
    \includegraphics[width=1\textwidth]{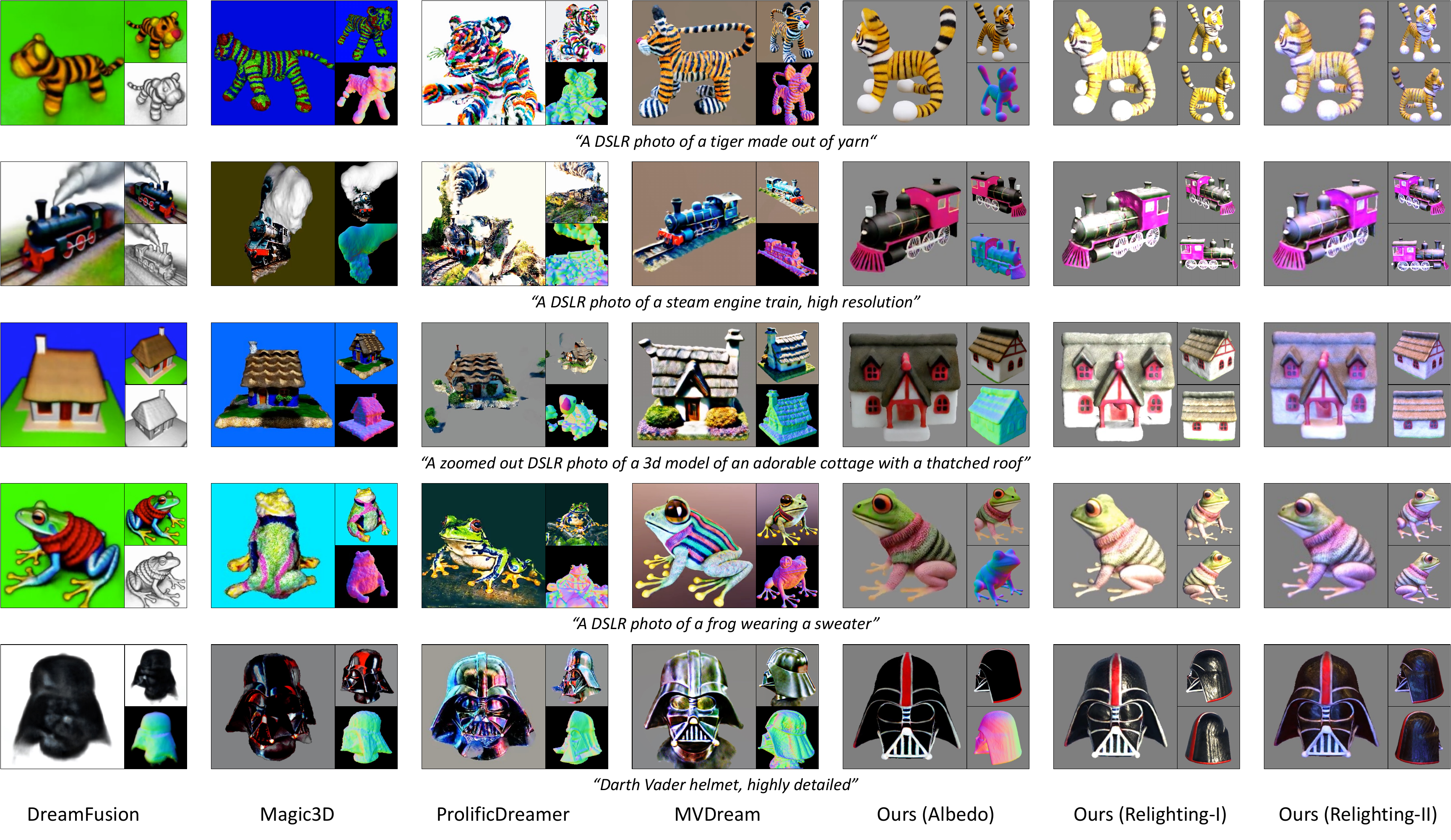}
    \vspace{-2em}
    \captionof{figure}{Comparison with baselines. UniDream presents clear albedo textures, completely smooth surfaces, and advanced relighting capabilities. The `Albedo' column demonstrates the albedo and normal properties of the 3D objects generated using our method. Meanwhile, the `Relighting-I' and `Relighting-II' columns demonstrate the effect of relighting on the generated PBR materials under white and purple lighting conditions, respectively.}
    \vspace{-3em}
    \label{fig:fig1}
\end{figure*}
{\let\thefootnote\relax\footnote{{* Equal contributions}}}
{\let\thefootnote\relax\footnote{{$\dagger$ Corresponding author}}}
\begin{abstract}
   Recent advancements in text-to-3D generation technology have significantly advanced the conversion of textual descriptions into imaginative well-geometrical and finely textured 3D objects. Despite these developments, a prevalent limitation arises from the use of RGB data in diffusion or reconstruction models, which often results in models with inherent lighting and shadows effects that detract from their realism, thereby limiting their usability in applications that demand accurate relighting capabilities. 
   To bridge this gap, we present UniDream, a text-to-3D generation framework by incorporating unified diffusion priors.
   Our approach consists of three main components: (1) a dual-phase training process to get albedo-normal aligned multi-view diffusion and reconstruction models, (2) a progressive generation procedure for geometry and albedo-textures based on Score Distillation Sample (SDS) using the trained reconstruction and diffusion models, and (3) an innovative application of SDS for finalizing PBR generation while keeping a fixed albedo based on Stable Diffusion model. 
   Extensive evaluations demonstrate that UniDream surpasses existing methods in generating 3D objects with clearer albedo textures, smoother surfaces, enhanced realism, and superior relighting capabilities. The project homepage is at: \href{https://yg256li.github.io/UniDream/}{https://UniDream.github.io}.
\end{abstract}
    
\section{Introduction}
The creation of high-quality 3D content, characterized by intricate geometric and textural details, holds important applications in various domains, including gaming, AR/VR, and artistic content creation. 
However, these applications generally require generated 3D objects to be relightable under particular lighting conditions, which is essential for their realism.
The current 3D models production methods that can meet these application requirements mainly rely on 3D artists, which brings a huge workload.

Recent methods~\cite{poole2022dreamfusion,lin2023magic3d,metzer2023latent,chen2023fantasia3d,wang2023prolificdreamer,yu2023text} have been exploring the generation of 3D assets from textual descriptions under the supervision of 2D diffusion models~\cite{rombach2022high,saharia2022photorealistic,zhuo2024lumina}.
For example, DreamFusion~\cite{poole2022dreamfusion} defines a learnable Neural Radiance Fields (NeRF)~\cite{mildenhall2021nerf} and optimizes it based on the Score Distillation Sampling (SDS).
To enhance generation quality, subsequent studies have diversified the pipeline, focusing on aspects like 3D representations~\cite{lin2023magic3d}, loss functions~\cite{wang2023prolificdreamer}, 3D prior~\cite{yu2023points}, and 2D diffusion models~\cite{shi2023MVDream, zhao2023efficientdreamer}.
Although these methods achieve impressive results, they cannot generate relightable objects, as they typically represent the underlying illumination and texture of an object as a holistic appearance, as shown in the first four columns in Fig.\ref{fig:fig1}, which results in inherent lighting and shadows baked into the texture of the generated object. 
When relighting, the inherent highlights and shadows on these textured surfaces can affect the realism of the object.

\begin{figure}
    \centering
    \vspace{-2em}
    \includegraphics[width=0.83\textwidth]{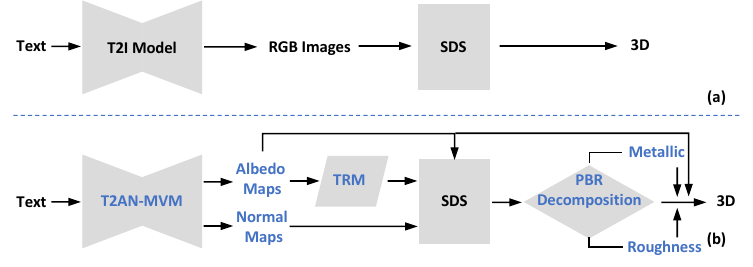}
    \vspace{-1.0em}
    \caption{Comparison of UniDream with other methods. (a) The existing RGB-based text-to-3D generation process; (b) UniDream's multi-stage generation process.}
    \vspace{-1em}
    \label{fig:method_comparison}
\end{figure}

In this paper, we present UniDream, a novel framework that allows generating relightable objects from textual descriptions.
Fig.\ref{fig:method_comparison} shows the fundamental difference between our method and other existing methods.
\textit{Our key idea is training a diffusion model that can provide both Physically-Based Rendering (PBR) material prior and multi-view geometry prior.}
Specifically, we first develop an albedo-normal aligned multi-view diffusion model (AN-MVM) for consistent multi-view image generation, which is trained on a set of paired albedo and normal data rendered from 3D object datasets.
Then, following the simplified Diesney BRDF model~\cite{burley2012physically}, we define a 3D representation that includes albedo, normal, roughness, and metallic properties, which are optimized based on the trained diffusion model and Stable Diffusion~\cite{rombach2022high} model.
Compared with previous text-to-3D methods~\cite{poole2022dreamfusion, lin2023magic3d, wang2023prolificdreamer, shi2023MVDream, yu2023text}, our approach is able to disentangle the illumination and PBR material, achieving high-quality relightable objects under different ambient lighting conditions, as shown in last three columns of Fig.\ref{fig:fig1}.

To robustly obtain 3D objects with PBR from 2D diffusion models, we have further developed a three-stage generation pipeline.
Initially, we utilize the albedo-normal aligned diffusion model (AN-MVM) to generate multi-view albedo and normal images.
Subsequently, we adapt a transformer-based reconstruction model (TRM) to convert multi-view albedo images to a 3D coarse model, and perform SDS refinement based on AN-MVM.
Finally, we fix the albedo and normal properties of the 3D model and optimize its roughness and metallic characteristics using the Stable Diffusion~\cite{rombach2022high} model.
Experiments in Sec.\ref{sec:exp}, demonstrate the superior capability of our three-stage pipeline in producing 3D objects with PBR property.

Overall, we propose a novel text-to-3D generation framework that can stably generate high-quality 3D objects through a multi-stage generation strategy utilizing SDS refinement based on multi-view diffusion and reconstruction models.
Extensive experiments have demonstrated UniDream's superiority in three critical areas:
(1) Realistic Materials: By disentangling lighting from textures, UniDream accurately generates PBR materials that approximate real-world textures and can be relit in various lighting conditions, greatly enhancing realism.
(2) Complete Geometry: Incorporating normal supervision into our optimization process, UniDream excels at generating more comprehensive geometric details than other existing methods, leading to more geometrically complete 3D objects.
(3) Stable Generation: Due to the introduction of 3D prior from reconstruction model and normal supervision in SDS process, UniDream's effectiveness in generating 3D objects is ahead of other methods.
\section{Related Works}
\subsection{Text-to-3D Generation}
In recent years, the field of text-to-3D content generation has seen significant advancements, largely inspired by the advancements in text-to-image generation. 
These advances have particularly been driven by methods employing CLIP-based guidance or score distillation.
Methods like ~\cite{jain2022zero,xu2023dream3d,mohammad2022clip} utilize the text-image relationship inherent in vision-language pre-trained models~\cite{MIR-2022-06-193, MIR-2022-07-224}, such as CLIP~\cite{radford2021learning} to facilitate general text-to-3D content creation.
Another innovative approach, pioneered by DreamFusion~\cite{poole2022dreamfusion}, employs score distillation to enhance the robustness of text-to-3D generation, which also has been further developed and expanded in subsequent works~\cite{lin2023magic3d, wang2023prolificdreamer,yu2023text,tang2023dreamgaussian}.
The central to both of these paradigms is the use of pre-trained text-to-image diffusion models as a foundation, enabling the creation of diverse and imaginative 3D content.
Furthermore, recent advancements have been developed by works~\cite{zhao2023efficientdreamer,shi2023MVDream,liu2023syncdreamer,liu2024sherpa3d}, which innovatively employ models derived from the Objaverse~\cite{deitke2022objaverse} 3D dataset. These models are used to render 2D RGB images, aiming to train text-to-multi-view image diffusion models. This approach represents a significant stride in optimization-based 3D model generation, as it enables the simultaneous generation of multiple coherent images. Such a technique effectively addresses the challenge of inconsistent directions in geometric optimization, thereby enhancing the consistency and quality of 3D model generation.

In a different paradigm, some methods have shifted towards training diffusion or reconstruction models directly on paired text-3D data. This strategy enables the creation of 3D models that inherently possess text-like semantics. 
A variety of techniques~\cite{cheng2023sdfusion,zheng2023locally,hui2022neural,nichol2022point,jun2023shap,gupta20233dgen,he2024gvgen} fall under this category of 3D diffusion generation. In these models, textual information serves as a conditional input, guiding the generation process. 
This approach emphasizes the manipulation of the underlying 3D data representation, ensuring that the resulting models are both semantically rich and accurate representations of the text descriptions.
Moreover, innovative strides have been taken in the realm of 3D reconstruction methods, particularly those grounded in transformer models, exemplified by~\cite{hong2023lrm, li2023instant3d, xu2024instantmesh, wang2024crm, xu2024grm,wei2024meshlrm,zhang2024gs,tochilkin2024triposr,zou2024triplane}. These methods introduce a novel perspective, enabling the generation of high-quality 3D models from text or images within seconds, courtesy of their efficient reconstruction networks. The adoption of 3D diffusion and reconstruction methodologies has gained prominence due to their impressive speed in generating 3D objects.

The 2D multi-view diffusion approach and the 3D reconstruction technique utilizing 3D data have provided substantial inspiration. UniDream deeply rethinks the essential principles of these methods and constructed with integrating the strengths of them.

\subsection{Materials Generation}
Estimating surface materials proposes a fundamental challenge in the field of computer vision. The Bidirectional Reflection Distribution Function (BRDF), as the predominant model, characterizes how the light is reflected off surfaces~\cite{nicodemus1965directional}. 
Early work focused on BRDF recovery concentrated on controlled lighting~\cite{aittala2013practical, nam2018practical}, yet they were less effective in real-world applications. 
However, recent advances in neural implicit methods~\cite{gao2019deep, srinivasan2021nerv, zhang2021nerfactor} have demonstrated potential in accurately estimating lighting and BRDF from image sets.
These methods utilize neural 3D reconstruction techniques to model complex lighting effects, simultaneously estimating shape, BRDF, and lighting. This achieves a more comprehensive decomposition of these elements. Nevertheless, the implicit representation of materials still poses limitations in their application. The recent advancement in differentiable rendering methods~\cite{munkberg2022extracting} addresses this issue by incorporating an explicit surface mesh optimization pipeline, allowing for the simultaneous estimation of BRDF and lighting.

Drawing inspiration from recent material estimation techniques, newer research has focused on generating surface materials for 3D objects. For instance, Fantasia3D~\cite{Chen_2023_ICCV} combines a physical differential rendering pipeline with SDS to produce detailed 3D objects with realistic surface materials. However, this approach sometimes mixes albedo with reflected light, resulting in blurred material property distinctions. To address this, MATLABER~\cite{xu2023matlaber} employs a latent BRDF auto-encoder, trained on an extensive dataset, to more effectively differentiate these properties. Building upon this, our method initiates with fixed albedo and normal generation and progressively incorporates other BRDF parameters, achieving a more natural and effective decomposition of surface material of 3D objects.
\section{Methodology}
\begin{figure*}[t]
\centering
\includegraphics[width=1\textwidth]{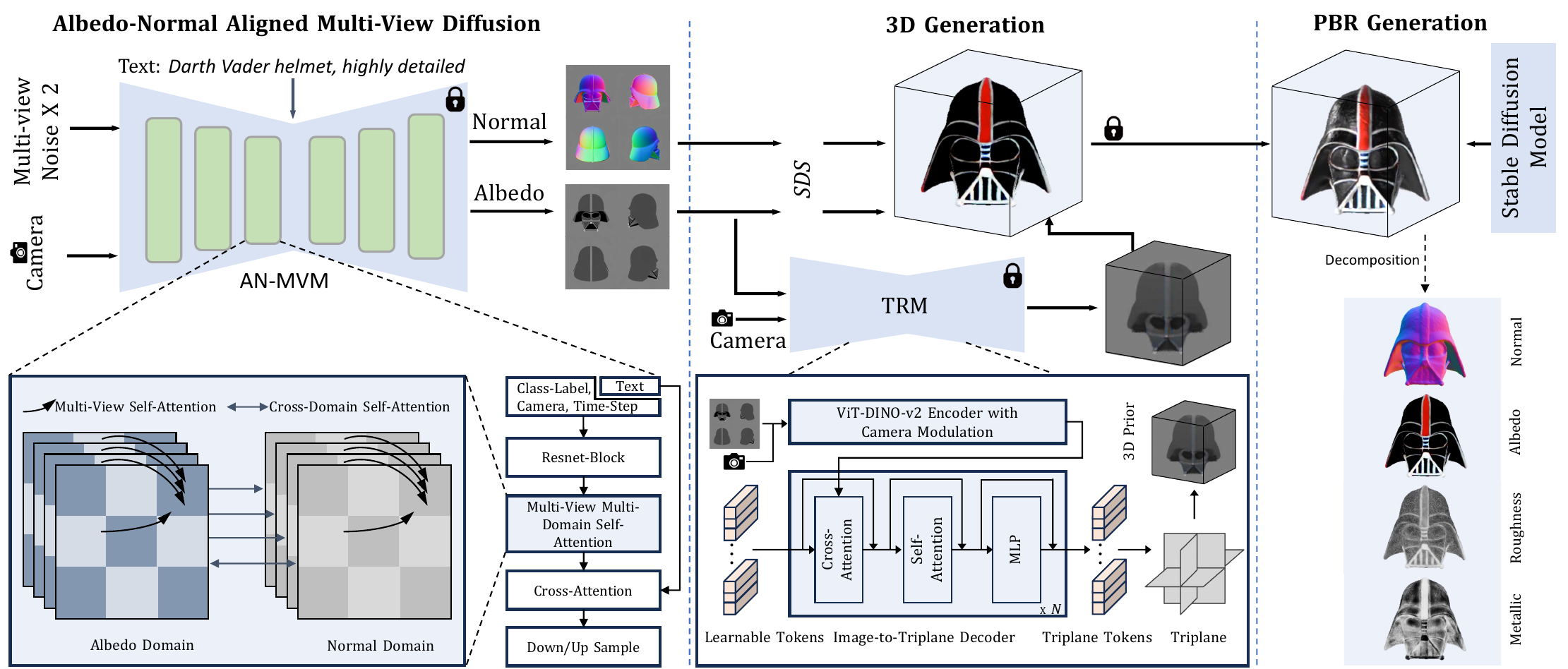}
% \vspace{0.1em}
\caption{Overview of UniDream. Left: the multi-view diffusion model generates multi-view images based on input text. Middle: first, four view albedo maps obtain 3D prior by the reconstruction model, and then the multi-view diffusion model performs SDS optimization based on the 3D prior to generate a 3D object with albedo texture. Right: using Stable Diffusion model to generate PBR material.}
% \vspace{-1em}
\label{fig:method}   
\end{figure*}
\noindent\textbf{Overview.}
As illustrated in Fig.\ref{fig:method}, UniDream can be structured into three stages and four modules. Firstly, upon receiving a text input, the pre-trained albedo-normal aligned multi-view diffusion model generates four view consistent albedo and normal maps (detailed in Section~\ref{method: multi-view diffusion}); secondly, these albedo maps are then fed into a transformer-based reconstruction model, which reconstructs a coarse 3D model to serve as an initial prior (as described in Section~\ref{method: reconstructor}); thirdly, building on this preliminary coarse model, SDS optimization is employed, using the albedo-normal aligned multi-view diffusion model to refine and produce a fine 3D object with detailed mesh and albedo texture (explained in Section~\ref{method: SDS refinement});
finally, we fix the albedo and geometric shapes, and then use a Stable Diffusion~\cite{rombach2022high} model to generate the corresponding materials (outlined in Section~\ref{method: PBR learning}).

\subsection{Albedo-Normal Aligned Multi-view Consistent Diffusion Model}\label{method: multi-view diffusion}
Our approach represents a departure from traditional methods that utilize RGB data to train diffusion models. We train an albedo-normal aligned text-to-multi-view diffusion model (AN-MVM) using albedo and normal maps rendered from 3D data. Based on the Stable Diffusion~\cite{rombach2022high} model framework, we perform multi-view and multi-domain diffusion modeling in the UNet module to establish multi-view consistency and multi-domain consistency. 
Specifically, within the UNet module design, we address three critical aspects: ensuring multi-view consistency, aligning the albedo with normal domains, and maintaining the semantic integrity of information from text to generated images.

~\\
\noindent\textbf{Multi-view Self-Attention.}
To ensure robust generalization, our AN-MVM model expands upon the capabilities of the pre-trained Stable Diffusion~\cite{rombach2022high} model by adapting it for multi-view training. This process, inspired by MVDream~\cite{shi2023MVDream}, initiates with the randomly selected four orthogonal views $x \in \mathbb{R}^{N \times H \times W \times C} $from the rendered multi-view dataset. We then encode the camera parameters $c \in \mathbb{R}^{N\times12}$ of these views using a two-layer MLP network. This procedure generates feature vectors $F_c$ with the same dimensions as time-step. These camera features are then added to the time-step features $F_t$, facilitating effective modulation of variance across different views.

In the architecture of our UNet module, we consolidate multi-view data within an additional dimension and perform self-attention mechanism between multiple views just before the cross-attention layer. This strategic design enables mutual constraints among the various multi-view inputs, effectively reinforcing consistency across multiple views during the diffusion process.
~\\

\noindent\textbf{Multi-Domain Self-Attention.}
Based on multi-view consistency and sharing a similar perspective to recent work~\cite{long2023wonder3d}, we further introduce multi-domain consistency.
Specifically, we introduce a distinct class label $L$ for the normal domain, and use a two-layer multi-layer perceptron (MLP) to encode this class label $L$ to obtain feature $F_l$ with the same dimension with the time-step features $F_t$, and $F_l$ is added to $F_t$ to control the generation process within the normal domain. Subsequently, we apply the self-attention mechanism to the corresponding views between the albedo and normal domains to ensure domain consistency.

It is important to highlight that achieving multi-view consistency in normal maps is notably straightforward, primarily due to the simplicity of their semantic content and the consistency of values at identical positions across various views in the world coordinate system. This inherent consistency in normal maps significantly facilitates the convergence process. Furthermore, the constraints we apply between the albedo and normal maps contribute to a quickly convergence in controlling multi-view albedo, streamlining the overall generation process.

~\\
\noindent\textbf{Text and Image Semantic Alignment.}
In order to solve the problem of potential semantic generalization loss caused by only using less 3D synthetic data during AN-MVM training, we use joint training to combine the 2D LAION-Aesthetics data with the published 3D data. Specifically, in line with the approach used by MVDream~\cite{shi2023MVDream} during our training process, each batch is randomly composed by 3D data or 2D data based on a probability distribution: there's a 70\% chance of using 3D data and a 30\% chance of incorporating 2D LAION-Aesthetics data. In this setting, while the normal domain is distinguished by class label, the albedo and 2D LAION-Aesthetics data are not. Moreover, to further differentiate between 3D and 2D data, we add the phrase ", 3D asset" to the captions of 3D data. This strategic inclusion of a significant proportion of 2D data plays a crucial role in enhancing the semantic alignment between the generated image content and the corresponding input text.

\subsection{Transformer-Based Reconstruction Model}\label{method: reconstructor}

Inspired by LRM~\cite{hong2023lrm} and Instant3D~\cite{li2023instant3d}, we have integrated reconstruction models into our text-to-3D generation pipeline, aiming to provide an initial 3D prior for enhancing text-to-3D generation performance. 
As illustrated in the TRM module of Fig.\ref{fig:method}, for each object in the AN-MVM training dataset, we randomly select four views $I_i$, identical elevation but orthogonal views.
Along with these views, the corresponding camera parameters, $C_i$, are used as inputs to the model.
In this framework, the pre-trained ViT-based DINO-v2~\cite{oquab2023dinov2} model, denoted as $F$ is employed to extract image features, $F_{i}$, from the albedo images of the four selected views.
Concurrently, a learnable camera modulation module processes the camera parameters for each view using $\text{MLP}^{\text{mod}}(C_i)$ and seamlessly integrates these encoded parameters into the image features.
Then we employ learnable tokens, denoted as $T_h$, as input to the transformer model. These tokens are designed to undergo cross-attention with the image features, allowing for an effective fusion of the input image information. Subsequently, this is followed by the integration of cross-attention and self-attention modules and a multi-layer perceptron ($\text{MLP}$) to form a transformer block. Multiple such transformer blocks work in sequence to decode the input tokens $T_h$ into refined triplane representations $T_h'$. The representations correspond to the semantic information of the input image. Finally, we decode these triplane representations using a $\text{MLP}$ decoder of NeRF to reconstruct the final 3D model, denoted as $\text{Gen}_{\text{3D}}$. This entire process is detailed in Eq.\ref{eq:recon}. And $M$ is the number of transformer layers. 
% \vspace{1.0em}
\begin{equation}
\label{eq:recon}
\begin{split}
&F_i = \text{MLP}^{\text{mod}}(C_i) \otimes F(I_{i}) \\
&T_h'= \text{MLP}(\text{SelfAttn}(\text{CrossAttn}(T_h, F_i))) \times M \\
&\text{Gen}_{\text{3D}} \Leftarrow \text{MLP}(T_h')
\end{split}
\end{equation}
% \vspace{0.5em}
Our approach differs from the configurations used in Instant3D in several key aspects to better adapt our model's requirements. Firstly, we remove the intrinsic camera parameters of the input multi-view images and only normalize and encode the extrinsic parameters to adapt the multi-view images output by AN-MVM. 
Furthermore, we use albedo instead of RGB for training to prevent the impact of lighting and shadows in RGB images on the triplane-NeRF reconstruction results. 
Additionally, in order to increase the resolution of reconstructed model and save training cost, we resize the reference views between $128\times128$ and $256\times256$ resolution, and randomly crop $128\times128$ images from the resized views to supervise the aligned region of rendered images.
Finally, in the supervision, we not only render albedo for supervision but also incorporate normal supervision to significantly accelerate the model's convergence speed and promote the geometry details.
These strategic enhancements enable our TRM to deliver superior reconstruction results while reducing the training cost, which will effectively increase the effectiveness of subsequent modules. 

\subsection{Score Distillation Sample (SDS) Refinement}\label{method: SDS refinement}
After acquiring the triplane-NeRF representation of the 3D model from TRM, we further refine it using our AN-MVM in conjunction with score distillation sample (SDS) to obtain higher quality 3D results. 

% \noindent\textbf{NeRF Refinement with SDS.}
When given a text input, a cascaded inference using AN-MVM and TRM produces a 3D coarse model represented by a triplane-NeRF $x = g(\theta)$, where $\theta$ is MLP network of NeRF, $g(\cdot)$ is the renderer, and $x$ is the generated view at a given camera pose. Subsequently, we employ AN-MVM with SDS to refine the 3D coarse model.
In details, the albedo and normal maps of four orthogonal views $x_{anmv}$ are rendered from the coarse model each iteration. After adding noise ${\epsilon}_{\text{anmv}}$, the frozen albedo-normal aligned multi-view diffusion model $\phi_{AN-MVM}$ is used to predict the noise $\hat{\epsilon}_{\phi_\text{{AN-MVM}}}(x_{\text{anmv},t};y,t)$ for all views across both domains simultaneously, where $t$ is the time-step representing noisy level, $y$ is the text condition, and $x_{\text{anmv},t}$ is the noised image. 
Subtracting the predicted noise, $\hat{\epsilon} - \epsilon$, offers a signal for aligning the rendered view $x_{anmv}$ with the text input $y$, as perceived by the AN-MVM.
UniDream updates the NeRF's MLP parameters by backpropagating the gradient through the rendering process using Score Distillation Sampling (SDS), as depicted in Eq.\ref{eq:SDS}.
% \vspace{0.5em}
\begin{equation}
\label{eq:SDS}
\begin{split}
    &\nabla_\theta{L_{\text{SDS}}({\phi}_{\text{AN-MVM}},g(\theta))} = \\
    &\mathbb{E}_{t, \epsilon}\! \! \left[ w(t)(\hat{\epsilon}_{\phi_{\text{AN-MVM}}}(x_{\text{anmv},t};y,t) - {\epsilon}_{\text{anmv}})\frac{\partial_x}{\partial_\theta} \right]
\end{split}
\end{equation}
% \vspace{-0.1em}
Here, $w(t)$ is a weighting function that depends on the timestep $t$. 
When calculating the final loss, we use weights of $0.8$ and $0.2$ for the two domains of albedo and normal respectively for weighted summation to ensure that fine geometry can be quickly optimized without neglecting the optimization of texture.

In order to get a better mesh, we adopt a strategy similar to Magic3D~\cite{lin2023magic3d}, incorporating DMTet~\cite{shen2021dmtet} refinement from NeRF representation to enhance mesh quality.

\subsection{Physically-Based Rendering (PBR) Material Generation} \label{method: PBR learning}
Different from RichDreamer~\cite{qiu2024richdreamer} generating route of PBR material, based on the geometry and albedo from the DMTet refinement, we employ the Stable Diffusion~\cite{rombach2022high} model to generate the PBR material.
We adopt the PBR material and illumination representations in Nvdiffrec~\cite{Munkberg_2022_CVPR}, which is recognized for its speed and efficiency.

For any 3D point, we predict its BRDF parameters, including the diffuse $k_d$, roughness $k_r$, and metalness $k_m$.
In our approach, following the DMTet refinement, where a hash grid and a multi-layer perceptron (MLP) are used to predict $k_d$, we introduce an additional hash grid and MLP to predict the roughness and metalness parameters, $k_r$ and $k_m$, respectively.
In accordance with Nvdiffrec~\cite{Munkberg_2022_CVPR}, we calculate the final rendering using:
\begin{equation}
\label{eq:pbr}
    L = k_d (1 - k_m) L_d(\omega_o, n) + k_s L_s(k_r, \omega_o, n)
\end{equation}
where $k_d (1 - k_m)L_d$ corresponds to the diffuse shading, while $k_s L_s$ means the specular shading. 
The terms $L_d$ and $L_s$ in the equation represent the diffuse and specular light components.
Please refer to \cite{karis2013real} for more details.

Due to the high quality performance of the previous stage, we fixed the albedo and normals. The model parameters are optimized based on the SDS loss.
To better accommodate the supervision from the Stable Diffusion model, we allow the ambient light to be optimized alongside the BRDF parameters, which is different from Fantasia3D~\cite{Chen_2023_ICCV} that fixes the lighting and MATLABER~\cite{xu2023matlaber} that uses a set of ambient light during training.
To avoid potential color interference that might arise from the Stable Diffusion model, we constrain the ambient light optimization to a single channel.
This channel only represents the magnitude of the lighting, effectively circumventing the introduction of any misleading color information.

\section{Experiments}\label{sec:exp}

\subsection{Implementation Details}
\noindent\textbf{Training Dataset Setup.}
Like most of previous work~\cite{shi2023MVDream, liu2023zero, liu2023syncdreamer}, we employed the public available 3D dataset Objaverse~\cite{deitke2022objaverse} for training. To enhance data quality, we implement a series of filtering rules as follows: \textit{no texture map; not a single object; accounting for less than 10\% of the picture; low quality; no caption information is provided in Cap3D~\cite{luo2023scalable}.}
After filtering, approximately 300K object remain, and then we follow the MVDream~\cite{shi2023MVDream} method to render multi-view albedo and normal data.

~\\
\noindent\textbf{Multi-View Diffusion Model Training Details.}
We follow Tune-A-Video~\cite{wu2023tune} to implement our multi-view diffusion model. During training, we use 32 A800 GPUs with $256\times256$ image resolution and a per-GPU image batch size of 128 (16 objects $\times$ 2 domains $\times$ 4 views) to train 50k iterations, which takes about 19 hours. In addition, the learning rate used is $1\times 10^{-4}$, and 10 times the learning rate is used for camera encoder's parameters.

~\\
\noindent\textbf{3D Reconstruction Model Training Details.}
We use random four views of 256$\times$256 images as input and produce $1,300 \times 768$ image features. The learnable tokens are a sequence of $(3\times32\times32) \times 512$. The image-to-triplane decoder are of 10 layers transformer with hidden dimentions 512.
We train the reconstruction model on 32 A800 GPUs with batch size 96 for 70,000 steps, taking about 3 days to complete. We set the coefficient $\lambda=2.0$ for $L_{lpips}$ and use the AdamW optimizer with a weight decay of 0.05 to train our model. And we use a peak learning rate of $4\times 10^{-4}$ with a linear 3K steps warm-up and a cosine scheduler.
\begin{figure*}[t]
% \begin{stride}
    \centering
    % \vspace{-1em}
    \centering
    \includegraphics[width=1\textwidth]{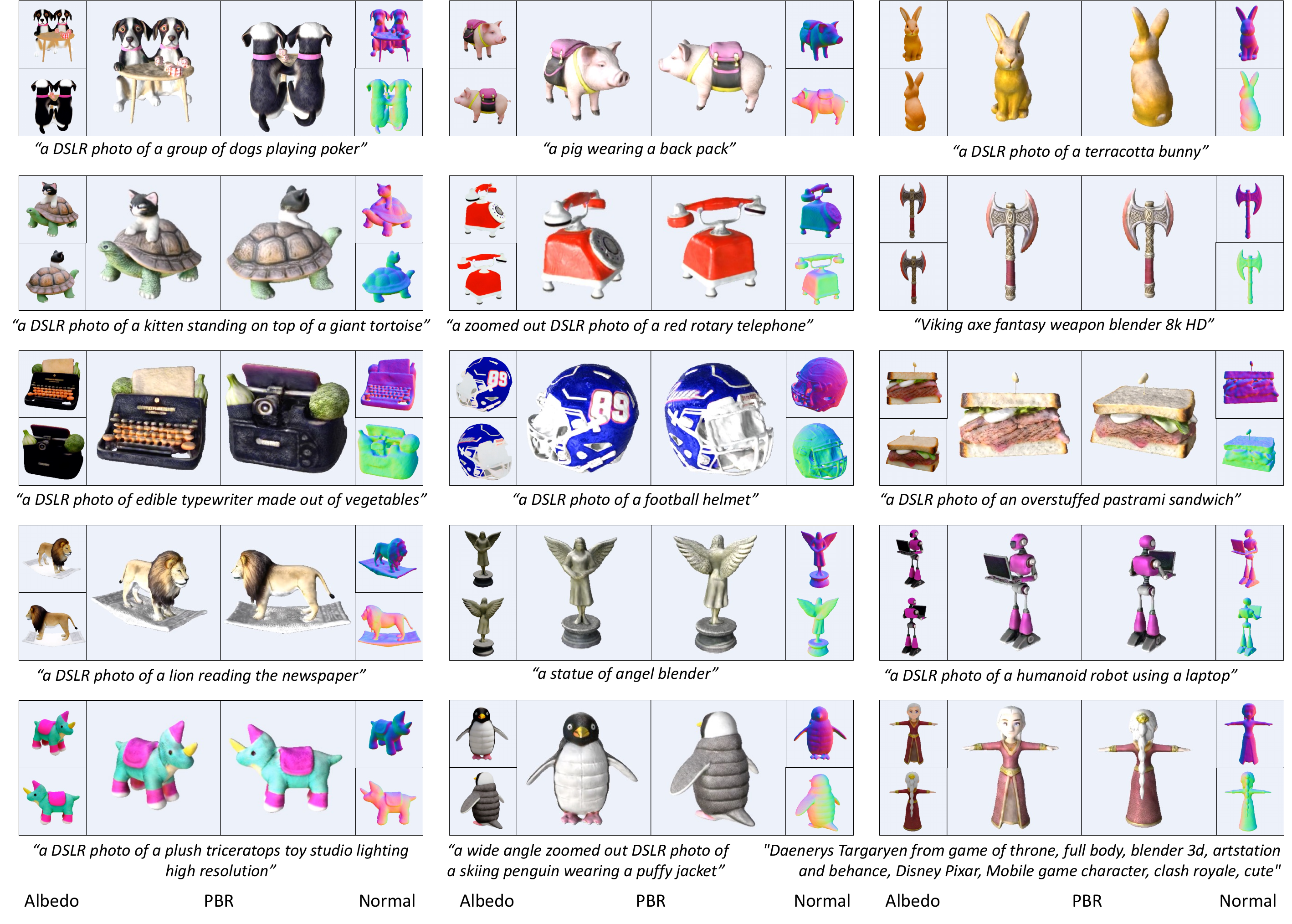}
    \vspace{-1.5em}
    \captionof{figure}{Illustrative overview of our method’s capabilities. We demonstrate the performance of 3D objects generated by our method in three dimensions: albedo, PBR, and normal.}
    \label{fig:show_case}
    \vspace{-1.em}
% \end{stride}
\end{figure*}
~\\

\noindent\textbf{Score Distillation Sample(SDS) Refining Details.}
We implement the refinement stages of NeRF and DMTet based on the Threestudio\footnote{https://github.com/threestudio-project/threestudio}. Specifically, we render four-view albedo and normal maps at the same time for SDS training. In the NeRF and DMTet refinement stages, we train 5,000 and 2,000 iterations respectively. During training, the loss weights of the albedo and normal domains are 0.8 and 0.2, and an `unsharp' operation~\cite{unsharp} is used in the last 500 iterations of each stage.

~\\
\noindent\textbf{PBR Material Generation Details.}
In the PBR material generation stage, the texture hash grid, derived from DMTet refinement, is repurposed and duplicated for the isolated learning of the parameters $\{k_r, k_m\}$, while maintaining a fixed parameter for the albedo texture. A new MLP is initialized for $\{k_r, k_m\}$ learning. The value ranges of these parameters are constrained within $[0.0, 0.9]$ for $k_r$ and $[0.08, 0.9]$ for $k_m$ to prevent erroneous PBR properties. We set the learning rate for the hash grid to $1\times10^{-4}$, while that of MLP is $0.1$. Additionally, the learning rate for the environment map is also set to $0.01$, coupled with total variation regularization. The environment map commences from an initialized studio lighting High-Dynamic-Range Imaging (HDRI) map. Image rendering resolution is 512$\times$512, and the model is trained for 2,000 iterations.
\subsection{Qualitative Comparisons}
We present representative results of UniDream in Fig.\ref{fig:show_case}, showcasing the albedo, PBR, and normal maps of the generated 3D objects. The text-to-3D objects created by UniDream, exhibit more complete and smoother geometric surfaces, clearer albedo in texture color distribution, and more realistic lighting effects. These features represent significant advancements over many previous methods.

In Fig.\ref{fig:fig1}, we compare the results generated by DreamFusion~\cite{poole2022dreamfusion}, Magic3D~\cite{lin2023magic3d}, ProlificDreamer~\cite{wang2023prolificdreamer}, MVDream~\cite{shi2023MVDream}, and UniDream.
We utilized the results from DreamFusion's official website for its first four cases. For the other cases, including those from Magic3D and ProlificDreamer, we employed Threestudio's implementation to acquire the results. The comparison reveals that UniDream produces semantically clearer 3D geometries and does not exhibit the 'Janus problem'.
In comparison with the methods, especially MVDream~\cite{shi2023MVDream}, illustrated in Fig.\ref{fig:fig1}, UniDream demonstrates a more complete and smoother geometric surface, attributed to the implementation of normal supervision. Moreover, an analysis of the last three columns in Fig.\ref{fig:fig1} reveals that UniDream possesses unique capabilities not typically found in existing methods. These include the ability to disentangle lighting and texture, exhibit relighting effects under various lighting conditions, and enhance the realism of the generated 3D objects.

\subsection{Quantitative Evaluations }
We conducted a quantitative evaluation of text-to-3D generation quality using CLIP Score~\cite{hessel2022clipscore, radford2021learning} and CLIP R-Precision~\cite{park2021benchmark} following methodologies from Dream Fileds~\cite{jain2022zero}, DreamFusion~\cite{poole2022dreamfusion}, and Cap3D~\cite{luo2023scalable}. 
Specifically, we generated 3D objects using 68 different prompts sourced from the DreamFusion and MVDream websites, employing DreamFusion, Magic3D, MVDream, and our UniDream. For the evaluation, four views (front, back, left, right) of each generated 3D object were rendered. We extracted text and image features using the CLIP ViT-B/32 model~\cite{radford2021learning} and calculated the CLIP score by averaging the similarity between each view and the corresponding text prompt. The detailed results, presented in Tab.\ref{tab:clip-score}, demonstrate that UniDream significantly surpasses DreamFusion, Magic3D, and MVDream in terms of CLIP Score and CLIP R-Precision. This indicates that UniDream is more effective at producing 3D results that are consistent with the text prompts.

Additionally, we performed a user study evaluating 68 results generated by each method, focusing on geometric texture quality and realism to discern the visual quality differences among the methods. Involving 22 participants, UniDream was distinguished as the preferred choice, securing 50.3\% of the votes. This outcome highlights the superior overall quality of our approach.

\begin{table}[t]
    \caption{Quantitatively compare UniDream with text-to-3D baseline methods by CLIP Score, CLIP R-Precision and user study.}
    \centering
    \scalebox{1}{
    \begin{tabular}{l|c|c|ccc} 
    \toprule 
    Methods  & User study&CLIP  & \multicolumn{3}{c}{CLIP R-Precision $(\%)\uparrow$} \\
               &(\%) $\uparrow$ & Score $(\uparrow)$ & R@1 & R@5 & R@10 \\
    \midrule
    DreamFusion~\cite{poole2022dreamfusion}  &  7.1 &71.0 &54.2 & 82.2 &91.5 \\
    Magic3D~\cite{lin2023magic3d}  &10.5  &75.1 &75.9 &93.5 &96.6 \\
    MVDream~\cite{shi2023MVDream} &32.1&75.7 &76.8 &94.3 &96.9 \\
    Ours  & \textbf{50.3}&\textbf{77.9} & \textbf{80.3} &\textbf{97.4} &\textbf{98.5} \\
    \bottomrule 
    \end{tabular}}
    \label{tab:clip-score}
    \vspace{-1em}
\end{table}

\begin{figure}
    \centering
    \vspace{-1em}
    \centering
    \includegraphics[width=1\textwidth]{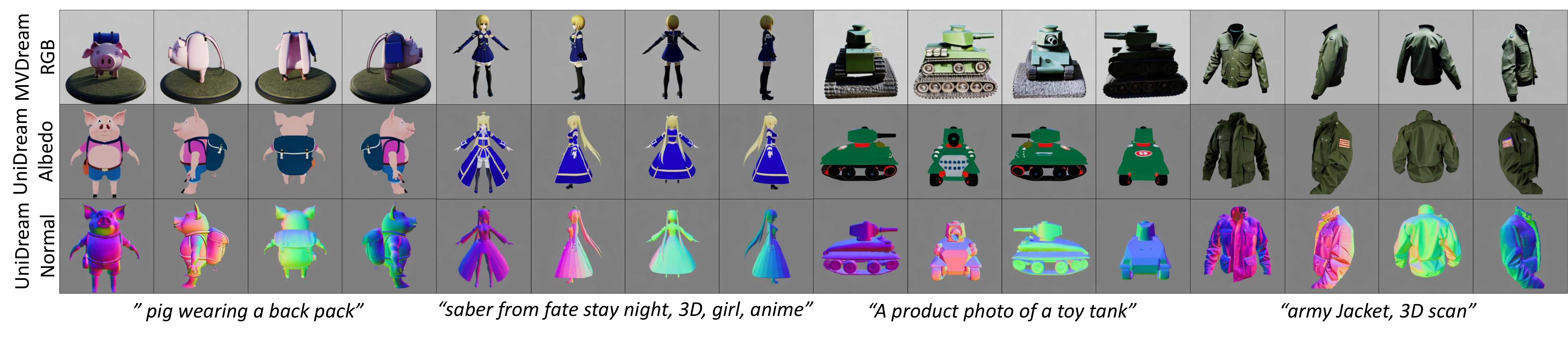}
    \vspace{-1.5em}
    \captionof{figure}{Comparison of multi-view results generated by MVDream (Row-1) and UniDream (Row-2-3).}
    \label{fig:unidream-mvdream-mv}
    \vspace{-2em}
\end{figure}

\begin{figure}[t]
    \centering
    % \vspace{-1em}
    \centering
    \includegraphics[width=1\textwidth]{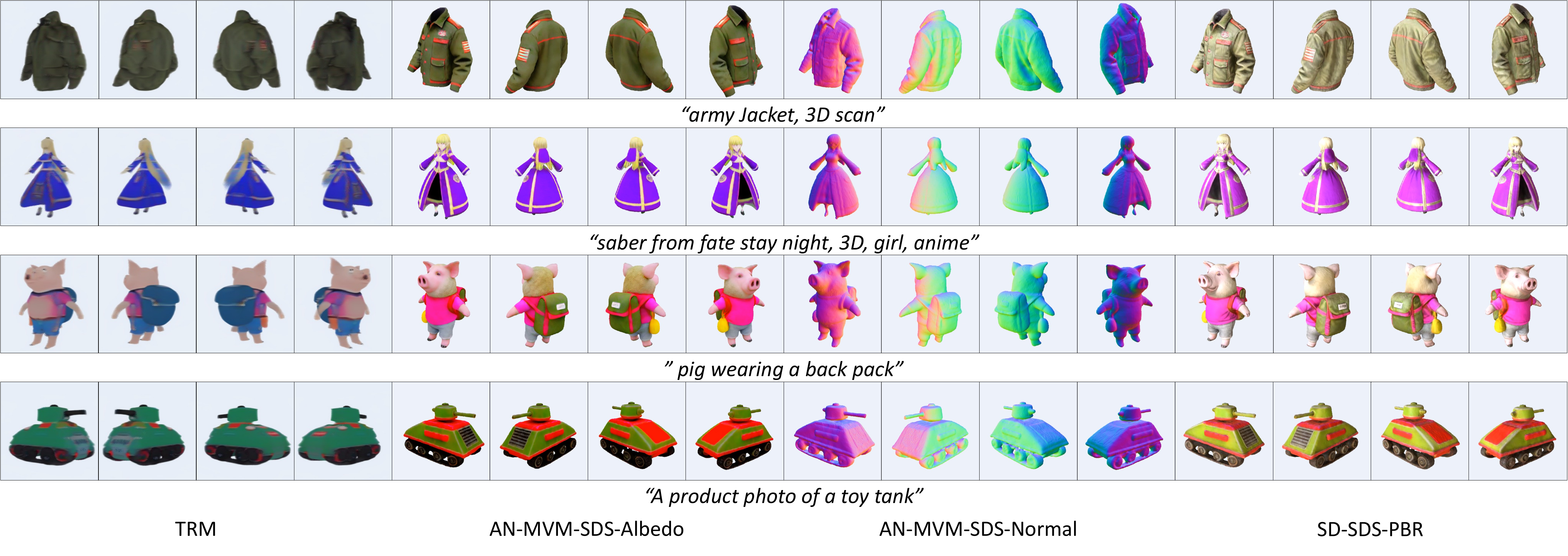}
    % \vspace{-2em}
    \captionof{figure}{Intermediate results of the UniDream generation process. From left to right: the results of TRM reconstruction, the albedo results and the normal results via SDS refinement based on the TRM result using the AN-MVM model, and the PBR result vis SDS generation based on the albedo and normal results using Stable Diffusion model.}
    \label{fig:unidream-trm}
    \vspace{-1.5em}
\end{figure}

\section{Ablation and Analysis}
\noindent\textbf{Comparison of Multi-view Diffusion Models.}
Generating multi-view images is a fundamental aspect of 3D content creation. To evaluate this, we compared the multi-view results produced by UniDream's AN-MVM with those of MVDream's multi-view diffusion model. For MVDream, we utilized the `sd-v2.1-base-4view' model based on its open-source inference code\footnote{https://github.com/bytedance/MVDream}. In addition, the same negative prompt used in MVDream's SDS optimization was applied for the multi-view inference in both MVDream and UniDream. As depicted in Fig.\ref{fig:unidream-mvdream-mv}, UniDream successfully maintains light and texture disentangling in its 2D output and produces normal maps with impressive consistency.

~\\
\noindent\textbf{Step-by-Step Comparison of Visualization During 3D Generation.}
By employing multiple views (as depicted in Fig.\ref{fig:unidream-mvdream-mv}) as input, UniDream's step-by-step 3D generation results are shown in Fig.\ref{fig:unidream-trm}. 
The left group shows the reconstruction results of TRM. Although the results are relatively rough, they still maintain fairly clear texture and geometric boundaries. The middle two groups present the generated albedo and normal results through SDS refinement using the AN-MVM model, based on the TRM results. After refinement, they exhibit a high quality of 3D models. The right group displays the generated PBR results by SDS optimization using the Stable Diffusion model with fixed albedo and normals, showcasing realistic 3D textures. The quality improves progressively from left to right, reflecting the excellent 3D generation capabilities of UniDream.

\begin{figure*}[h]
    \centering
    \vspace{1em}
    \centering
    \includegraphics[width=1\textwidth]{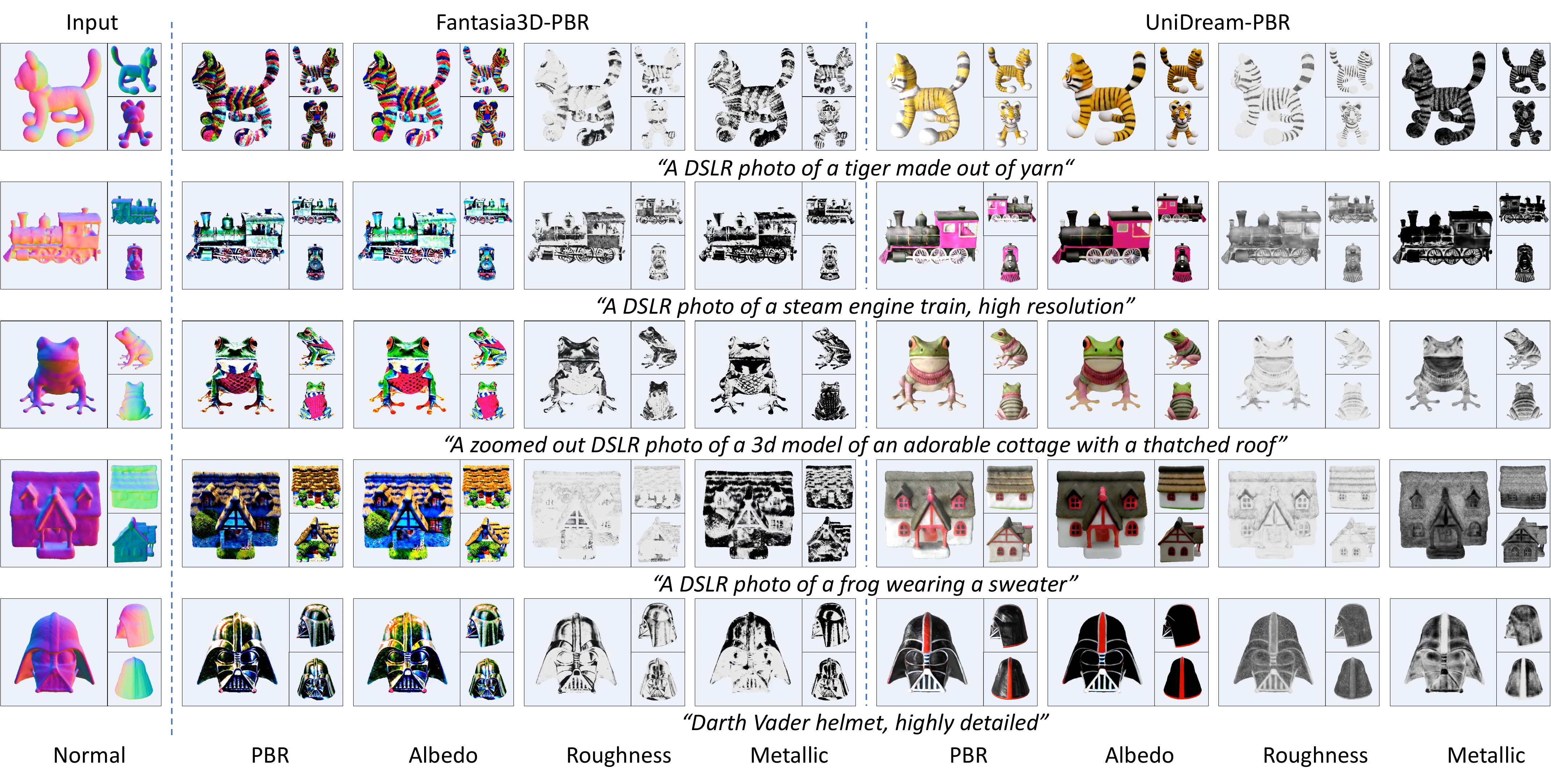}
    % \vspace{-1.5em}
    \captionof{figure}{Results comparison of generated PBR materials. Left: the same geometry input; Middle: the PBR material components generated by Fantasia3D; Right: the PBR material components generated by UniDream.}
    \label{fig:unidream-pbr-com}
    \vspace{1em}
\end{figure*}

~\\
\noindent\textbf{Comparison of Generated PBR Materials.}
As shown in Fig.\ref{fig:unidream-pbr-com}, we present the results of PBR material generation in comparison with Fantasia3D~\cite{chen2023fantasia3d}. It is evident that Fantasia3D struggles to disentangle lighting and textures, often resulting in lighting and shadows being baked into the appearance of 3D objects. Conversely, our method effectively disentangle lighting and textures, enabling the generation of realistic, relightable 3D objects.

\begin{figure}[t]
    \centering
    % \vspace{1em}
    \includegraphics[width=1\textwidth]{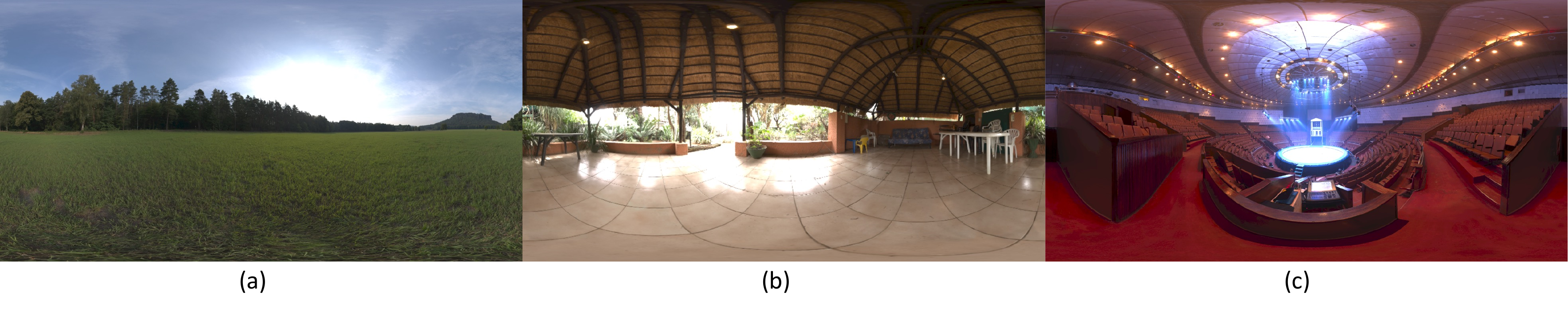}
    % \vspace{-2em}
    \captionof{figure}{Different environment maps used by UniDream.}
    \vspace{-2em}
    
    \label{fig:envir}
\end{figure}

~\\
\noindent\textbf{Relighting Comparison Driven by Environment Maps.} 
Fig.\ref{fig:envir} shows the different environment maps used by UniDream. Fig.\ref{fig:envir}(a) is the environment map used by Fig.\ref{fig:unidream-pbr-com} and the visualize of UniDream's overall capabilities as shown in Fig.\ref{fig:show_case}. Fig.\ref{fig:envir}(b) and Fig.\ref{fig:envir}(c) are the environment maps used by 'Relighting-I' and 'Relighting-II' in the teaser of UniDream respectively. Changing different environment maps will produce different rendering results, which reflects UniDream's excellent relightability.

\section{Conclusion}
In this paper, we propose for the first time a relightable text-to-3D generation paradigm, UniDream, which is based on an albedo and normal aligned multi-view diffusion model. Thanks to the disentangling of lighting and textures, the 3D models generated by our method can be relit, thereby enhancing their realism and usability. We provide a detailed discussion and analysis of each module in UniDream, and extensive results underscore the superiority of our approach.

~\\
\noindent\textbf{Limitations and future work.}
While UniDream demonstrates clear advantages from multiple perspectives, it also has certain limitations due to being trained on only approximately 300k Objaverse~\cite{deitke2022objaverse} data.
Primarily, there may be constraints in semantic generalization, leading to potential challenges with complex combinational concepts. Additionally, issues in material generalization could arise, such as in accurately simulating materials with transparent properties.
Our subsequent work will primarily concentrate on enhancing the generalization of the pipeline. Moreover, there is a critical demand to upgrade our rendering pipeline, aiming to boost the realism and visual fidelity of the generated 3D models. By incorporating path tracing, renowned for its realistic simulation of lighting and shadow effects, we anticipate a substantial improvement in rendering quality.

\par\vfill\par

\bibliographystyle{splncs04}
\bibliography{main}
\end{document}